\pdfoutput=1

\documentclass[11pt]{article}

\usepackage[final]{EMNLP2023}

\usepackage{times}
\usepackage{latexsym}

\usepackage[T1]{fontenc}

\usepackage[utf8]{inputenc}

\usepackage{microtype}

\usepackage{inconsolata}
\usepackage{graphicx}
\usepackage{amsmath,amssymb} 
\usepackage{color}
\usepackage{xcolor}
\usepackage{xspace}
\usepackage{booktabs}
\usepackage{enumitem}

\title{Show, Write, and Retrieve: Entity-aware Article Generation and Retrieval}

\author{Zhongping Zhang \qquad Yiwen Gu \qquad Bryan A. Plummer \\
  Boston University\\
  \texttt{ \{zpzhang, yiweng, bplum\}@bu.edu} 
  }

\begin{document}
\makeatletter
\DeclareRobustCommand\onedot{\futurelet\@let@token\@onedot}
\def\@onedot{\ifx\@let@token.\else.\null\fi\xspace}

\def\eg{\emph{e.g}\onedot} \def\Eg{\emph{E.g}\onedot}
\def\ie{\emph{i.e}\onedot}
\def\Ie{\emph{I.e}\onedot}
\def\cf{\emph{cf}\onedot} \def\Cf{\emph{Cf}\onedot}
\def\etc{\emph{etc}\onedot} \def\vs{\emph{vs}\onedot}
\def\wrt{w.r.t\onedot} \def\dof{d.o.f\onedot}
\def\iid{i.i.d\onedot} \def\wolog{w.l.o.g\onedot}
\def\etal{\emph{et al}\onedot}
\makeatother

\definecolor{lightmauve}{rgb}{0.86, 0.82, 1.0}
\definecolor{lightgoldenrodyellow}{rgb}{0.98, 0.98, 0.82}
\definecolor{lightapricot}{rgb}{0.99, 0.84, 0.69}
\definecolor{lightblue}{rgb}{0.55, 0.85, 0.9}
\definecolor{lightskyblue}{rgb}{0.53, 0.81, 0.98}
\definecolor{non-photoblue}{rgb}{0.64, 0.87, 0.93}
\definecolor{lightcornflowerblue}{rgb}{0.6, 0.81, 0.93}
\definecolor{lightgreen}{rgb}{0.56, 0.93, 0.56}
\definecolor{lightseagreen}{rgb}{0.13, 0.7, 0.67}
\definecolor{lightpink}{rgb}{1.0, 0.71, 0.76}

\maketitle
\begin{abstract}
Article comprehension is an important challenge in natural language processing with many applications such as article generation or image-to-article retrieval. Prior work typically encodes all tokens in articles uniformly using pretrained language models. However, in many applications, such as understanding news stories, these articles are based on real-world events and may reference many named entities that are difficult to accurately recognize and predict by language models. To address this challenge, we propose an ENtity-aware article GeneratIoN and rEtrieval (\textsc{Engine}) framework, to explicitly incorporate named entities into language models. \textsc{Engine} has two main components: a named-entity extraction module to extract named entities from both metadata and embedded images associated with articles, and an entity-aware mechanism that enhances the model's ability to recognize and predict entity names. We conducted experiments on three public datasets: GoodNews, VisualNews, and WikiText, where our results demonstrate that our model can boost both article generation and article retrieval performance, with a 4-5 perplexity improvement in article generation and a 3-4\% boost in recall@1 in article retrieval. We release our implementation at \href{https://github.com/Zhongping-Zhang/ENGINE}{this http URL}.
\end{abstract}

\section{Introduction}
\label{sec:introduction}

\begin{figure}[t]
    \centering
    \includegraphics[width=1.0\linewidth]{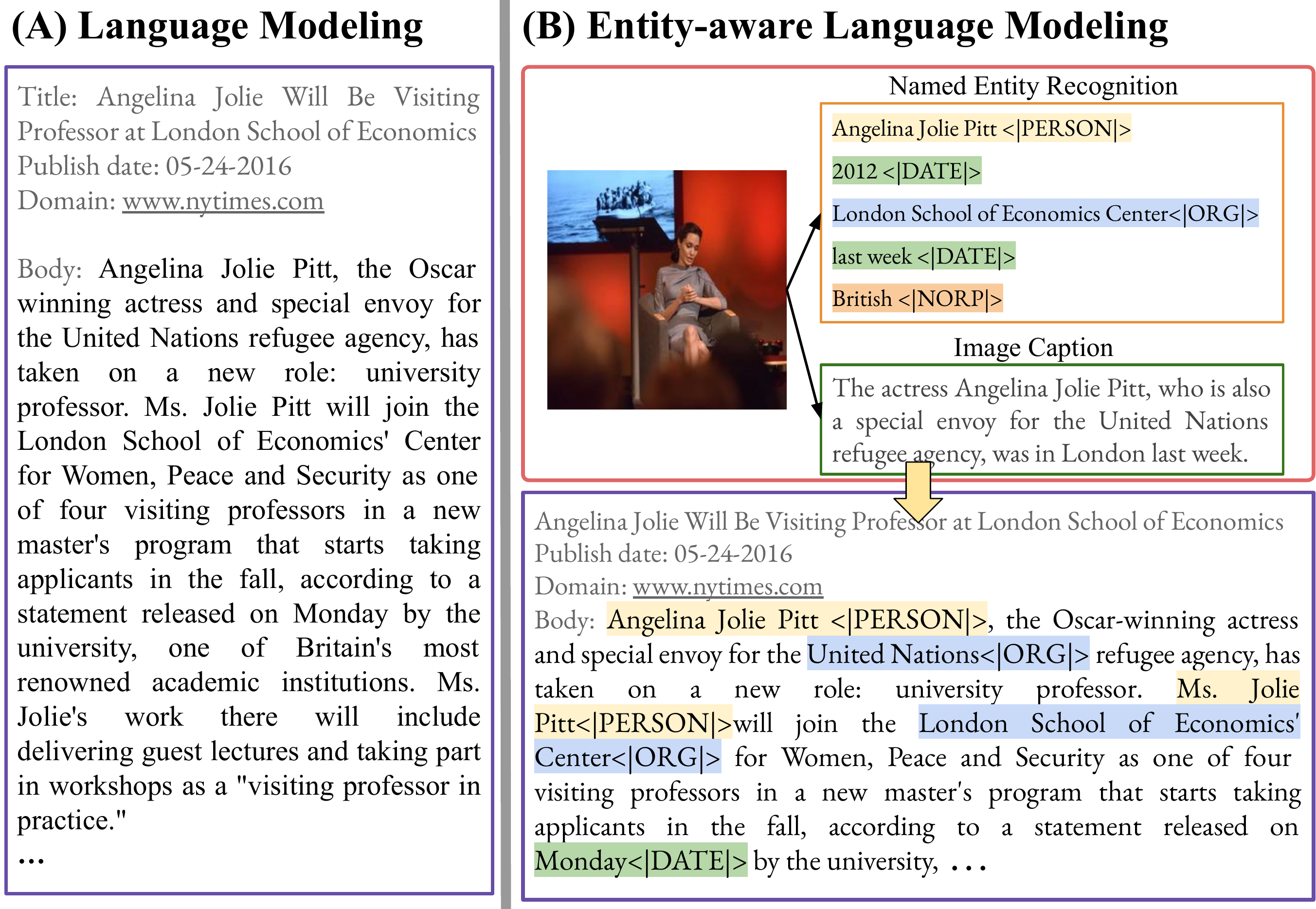}
    \caption{We propose entity-aware language modeling for article generation and retrieval. Prior work~\citep{brown2020language, zellers2020defending, ouyang2022training}, shown in (A), typically comprehends articles by uniformly encoding all text tokens. However, these approaches face challenges in accurately recognizing and predicting named entities. In our paper, shown in (B), we propose a method to extract named entities from embedded images and explicitly model the named entities in articles, boosting the performance of both article generation and article retrieval.}
    \label{fig:overview}
\end{figure}

Comprehending articles enables a wide range of applications such as story generation~\citep{fan2018hierarchical, peng2018towards}, image-to-text retrieval~\citep{tan2022newsstories}, automated journalism~\citep{leppanen2017data, brown2020language}, defending against misinformation~\citep{zellers2020defending, tan2020detecting}, and writing Wiki articles~\citep{banerjee2016wikiwrite, merity2016pointer}, among others. Inspired by the impressive capability of large language models, recent work (\eg,~\citet{radford2019language, brown2020language, gpt-j}) generate or retrieve articles by training language models on massive datasets (\eg, The Pile~\citep{gao2020pile} or LAION400M~\citep{schuhmann2021laion}). These models typically uniformly encode all text tokens in the articles including named entities~\citep{radford2018improving, brown2020language, zellers2020defending}. In other words, named entities like organizations, places, and dates are modeled together with other text, as illustrated in Figure \ref{fig:overview}(A). However, it poses a challenge for these models to accurately recognize and predict named entities, as they can be unique to specific articles. \Eg, in Figure \ref{fig:overview}(A), entities like ``Ms. Jolie'' and ``Jolie Pitt'' may only appear in articles related to the celebrity ``Angelina Jolie.''

Directly extracting named entities from user-provided prompts is a straightforward entity-aware approach used by methods addressing news image captioning~\citep{biten2019good, tran2020transform, liu2020visualnews}. However, these methods do not generalize well to article generation and retrieval since they rely on substantial contextual information (via the articles) as well as a direct indication of entities that may appear in predicted captions. For example, in Figure~\ref{fig:overview} (A), the article mentions that Angelina Jolie Pitt is an Oscar winner, but this information is not present in the other metadata like the image captions.  Thus, a language model generating an article must infer this named entity information rather than directly extracting it from the metadata.  Even if a list of named entities were provided, an article generation model must determine where and when to use each of them. In contrast, in news image captioning, entities used in the caption almost always appear in the body of the article~\citep{liu2020visualnews,tan2020detecting}, and the image itself will directly inform what named entities should be used for its caption. Thus, as we will show, adapting entity-aware mechanisms from related work (\eg,~\citep{liu2020visualnews,dong2021injecting}) results in poor performance in our task.

To address the aforementioned issues, we propose an \textbf{EN}tity-aware article \textbf{G}enerat\textbf{I}o\textbf{N} and r\textbf{E}trieval (\textsc{Engine}) framework to explicitly incorporate and model named entities in articles. \textsc{Engine} mainly consists two modules: a named-entity extraction module and an entity-aware module. In the named-entity extraction module, we show that providing a list of named entities for article generation improves performance. However, creating such lists does require a small overhead cost. Thus, we also demonstrate we can improve performance without manual input. 

As shown in Figure \ref{fig:overview} (B), we observe that images associated with an article often contain information about the article's events. Therefore, we explore leveraging large vision-language models to extract named entities from embedded images. Specifically, we employ CLIP~\citep{radford2021learning} to automatically select a set of likely named entities from embedded images. In the entity-aware module, we introduce special tokens after each entity name to indicate its entity category. In this case, \textsc{Engine} models the named entity and its entity category jointly. An additional benefit brought by our entity-aware mechanism is the named-entity recognition (NER) ability, \ie, our model not only recognizes and predicts the entity names but also predicts their entity categories simultaneously.

In summary, the contributions of this paper are:
\begin{itemize}[nosep,leftmargin=*]
    \item  We propose an entity-aware language model, \textsc{Engine}, for article generation and retrieval. Compared to existing language models~\citep{brown2020language, zellers2020defending, radford2021learning, sun2023eva}, our entity-aware mechanism enhances the recognition and prediction of named entities by jointly modeling entity names and their entity categories, boosting the performance of article generation and retrieval.
    \item We introduce a named-entity extraction method to recognize named entities from embedded images in articles, eliminating the overhead in manually creating a list of named entities that will appear in the articles. 
    \item Experiments on GoodNews~\citep{biten2019good} and VisualNews~\citep{liu2020visualnews} show a perplexity gain of 4-5 points for article generation and a Recall@1 boost of 3-4\% for article retrieval. We also show that \textsc{Engine} generalizes via zero-shot transfer to WikiText~\citep{merity2016pointer}.
    \item We perform comprehensive experiments on human evaluation and machine discrimination, validating that \textsc{Engine} produces more realistic articles compared to prior work~\citep{radford2019language, zellers2020defending}. This suggests that our model can potentially contribute additional training data for the development of more powerful machine-generated text detectors.
\end{itemize}

\begin{figure*}[t]
    \centering
    \includegraphics[width=1.0\linewidth]{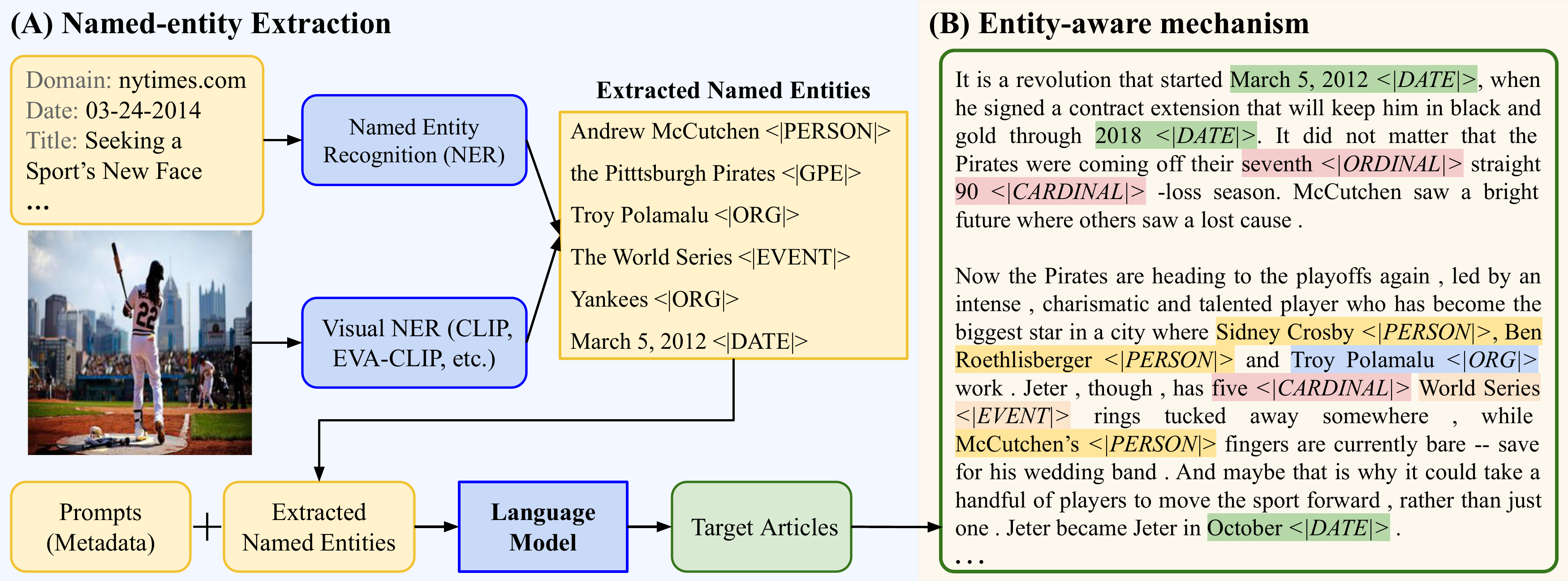}
    \caption{\textbf{\textsc{Engine} overview}. Our model mainly consists of two modules: (A) Named-entity Extraction: We extract named entities from both metadata and embedded images in articles. See Section \ref{sec:Engine_NE_extraction} for detailed discussion; (B) Entity-aware mechanism: Each named entity is associated by its corresponding entity category. \textsc{Engine} models the entity name and category jointly to avoid inconsistency between named entities in images and article text. See Section \ref{sec:Engine_NE_aware} for detailed discussion.}
    \label{fig:ENGINE_framework}
\end{figure*}

\section{Related Work}
\label{sec:related_work}
\noindent \textbf{Article Generation} in recent work uses large-scale pretrained transformer models that can be separated into unconditional text generation~\citep{radford2018improving, radford2019language} and conditional text generation~\citep{brown2020language, zellers2020defending}. Generating articles via unconditional samples has been found to be less effective, since the models may interpret the first sentence of articles as a tweet and start posting responses ~\citep{brown2020language}. To enable controllable generation, GPT3~\citep{brown2020language} conditions article generation on titles and the initial sentences of articles. Grover~\citep{zellers2020defending} decomposes news articles into distinct parts and conditions generation on metadata like the author or organization. In this paper, we further explore the impact of named entities and embedded article images. Specifically, \textsc{Engine} produces articles conditioned on both metadata and embedded article images, with a dedicated focus on the explicit extraction and modeling of named entities.

\noindent \textbf{Article Retrieval} is commonly accomplished by image-text matching frameworks. Early work on image-text matching has primarily focused on developing bespoke models~\citep{wang2017adversarial,gu2018look,nguyen2018improved,nam2017dual} with various retrieval loss functions, such as triplet loss~\citep{schroff2015facenet} or proxy anchor loss~\cite{kim2020proxy}. However, these models are often domain-specific and limited in the expressiveness of the text. In recent work, large vision-language models~\citep{radford2021learning,sun2023eva,li2022blip} have addressed these limitations by contrastive image-language pretraining on massive datasets (\eg, LAION-400M~\citep{schuhmann2021laion}). These large vision-language models have demonstrated exceptional performance in zero-shot image-text matching. Therefore, in our experiments, we employ the pretrained vision-language model EVA-CLIP~\cite{sun2023eva} to extract visual and text representations from articles, and compute the cosine similarity between these representations to obtain the retrieval predictions.  

\noindent \textbf{Entity-aware mechanisms} have been applied in closely related tasks, such as news image captioning, which aims to caption images based on articles and images. \citet{ramisa2017breakingnews} proposed an end-to-end framework that takes the concatenation of article and image features as input and outputs captions by an LSTM decoder. However, this approach often fails to predict named entities that were not seen during training. Thus, more recent work  boosts performance by extracting entity representations from user-provided articles and inserts them into generated templates~\citep{biten2019good, tran2020transform, liu2020visualnews}. In this paper, we effectively reverse the inputs and outputs of these approaches, \ie, we generate an article based on images and captions rather than generating captions based on images and articles. As discussed in the Introduction, this shift breaks the assumptions used by entity-aware mechanisms in image captioning, causing them to not generalize well in our article generation and retrieval task.

\section{\textsc{Engine}: ENtity-aware article GeneratIoN and rEtrieval}
\label{sec:Engine}
Given user-provided prompts and embedded article images, our task aims to more accurately represent articles by recognizing and predicting named entities. Thus, we incorporate candidate named entities as an additional control to our language model. We propose an entity-aware mechanism to jointly model entity names and their corresponding entity categories in Section \ref{sec:Engine_NE_aware}. To enrich the entity information available to language models, we introduce our named-entity extraction approach in Section \ref{sec:Engine_NE_extraction}. Finally, we introduce the learning strategy of \textsc{Engine} in Section \ref{sec:Engine_overall_strategy}. Figure \ref{fig:ENGINE_framework} provides an overview of our method.

\subsection{Entity-aware Mechanism}
\label{sec:Engine_NE_aware}
As discussed in the Introduction, accurately recognizing and predicting named entities can help avoid inconsistencies between the associated images and the textual content of an article. For example, in NBA news, an entity-aware model should be able to predict ``Curry'' given the preceding word ``Stephen,'' whereas traditional language models might struggle in this regard. Existing methods typically model named entities uniformly with other text, making the leverage of named entities less effective. To help our language model be aware of named entities, we insert the entity category predicted by SpaCy~\citep{honnibal2017spacy} after each entity name. We use special tokens as the indicator of these entity types. Then the entity name and its corresponding category are modeled jointly by \textsc{Engine}. We visualize our entity-aware mechanism in Figure \ref{fig:ENGINE_framework} (B).

\subsection{Named-entity Extraction}
\label{sec:Engine_NE_extraction}
Our entity-aware mechanism in Section \ref{sec:Engine_NE_aware} enhances a language model's ability to recognize and predict named entities. However, we find that the entity information extracted from the metadata may not be sufficient for article generation. Thus, as shown in Figure \ref{fig:ENGINE_framework} (A), we explore named entity extraction methods from various sources within articles, such as embedded images. Below we discuss two ways to create the named entity list.

\noindent \textbf{Oracle named-entities.} This approach assumes that we are provided with all the named entities that would appear in articles, \eg, named entities provided by a user. To simulate user-provided named entities, we extract named entities from news articles using SpaCy~\citep{honnibal2017spacy}. This list is then provided as input to our model.

\noindent \textbf{CLIP-based NER.} Existing Named Entity Recognition (NER) methods~\citep{yadav2019survey, li2020survey} primarily distinguish named entities within text documents and are not designed for NER involving images (referred to as Visual-NER in Figure \ref{fig:ENGINE_framework}(A)). However, we note that CLIP~\citep{radford2021learning} was trained on 400 million image-text pairs collected from the internet, many of which likely contain named entities. Thus, we use CLIP to build an open-ended Visual-NER framework. First, we construct a candidate list of named entities for each image by extracting entities from the articles in the dataset using SpaCy~\citep{honnibal2017spacy}. Subsequently, we use CLIP to predict the similarity between the article images and the candidate entities. The top $k$\footnote{We set $k$ to 10 in this paper.} entities are then provided as input to our model.

\subsection{Learning Strategy}
\label{sec:Engine_overall_strategy}
\noindent \textbf{Language Modeling.} Given a set of documents $\{x_1, x_2, ..., x_n\}$ each with variable length sequences of symbols $\{s_1, s_2, ..., s_m\}$, the statistical language model of a text document $x$ can be represented by the probability of next symbol given all the previous ones~\citep{bengio2003neural}:
\begin{align}
    p(x) = \prod_{i=1}^{m}p(s_i|s_{1}, ..., s_{i-1}),
\label{eq1: lm}
\end{align}
where each symbol $s_i$ is processed uniformly and the document $x$ is viewed as an unstructured \emph{text} field (also referred as \colorbox{lightmauve}{body} field later). Language models based only on Eq.\ \ref{eq1: lm} produce articles via unconditional samples. Thus, these models are not suitable for controllable generation~\citep{hu2017toward}. Instead, the language model can be formulated by the joint distribution of separate fields decomposed from the article $x$ ~\citep{zellers2020defending}:
\begin{align}
    p(x) = p({\rm \colorbox{lightapricot}{meta}, \colorbox{lightmauve}{body}}),
\label{eq2: lm_fields}
\end{align}
\noindent where \colorbox{lightapricot}{meta} is a data-dependent term consisting of a set of subfields. For instance, \colorbox{lightapricot}{meta} includes \emph{date}, \emph{title}, \emph{summary} in GoodNews~\citep{biten2019good} and \emph{domain}, \emph{date}, \emph{topic}, \emph{title} in VisualNews~\citep{liu2020visualnews}. Thus, we model $x$ by:
\begin{align}
    p(x) = p({\rm \colorbox{lightmauve}{body}|\colorbox{lightapricot}{meta}}) p({\rm \colorbox{lightapricot}{meta}}).
\label{eq3: lm_fields_order}
\end{align}
Based on Eq.\ \ref{eq3: lm_fields_order}, we further introduce special tokens $<$start-$\tau$$>$ and $<$end-$\tau$$>$ to indicate the boundaries of field $\tau$. The content of a target field $\tau$ is sampled from the model starting with $<$start-$\tau$$>$ and ending with $<$end-$\tau$$>$. Given the named entities extracted by our method (from Section \ref{sec:Engine_NE_extraction}), Eq.\ \ref{eq3: lm_fields_order} is re-formulated as:
\begin{align}
    p(x) = p(\colorbox{lightmauve}{body} | \colorbox{lightapricot}{meta}, \colorbox{lightpink}{entity})p(\colorbox{lightapricot}{meta}, \colorbox{lightpink}{entity}).
\label{eq4: lm_fields_visual}
\end{align}
To sample from Eq.\ \ref{eq4: lm_fields_visual}, we define a canonical order\footnote{We define canonical order in Goodnews~\citep{biten2019good} as: domain, date, named-entity, title, caption, summary, body; and Visualnews~\citep{liu2020visualnews} as: domain, date, topic, named-entity, title, caption, body. } among the fields (or subfields) of articles $\mathcal{F}: (f_1<f_2<...<f_{|\mathcal{F}|})$ and model the articles left-to-right in the order using Eq.\ref{eq1: lm}: $s^{f_1}_1, s^{f_2}_2, ... , s^{f_{|\mathcal{F}|}}_{|f_{|\mathcal{F}|}|}$. 

\noindent \textbf{Architecture.} Following~\citep{zellers2020defending},  \textsc{Engine} uses the GPT2 architecture~\citep{radford2019language} for article generation. We experiment with three model sizes: (1) \textsc{Engine}-Base has 12 layers and 124 million parameters, on par with GPT2-124M and \textsc{Grover}-Base; (2) \textsc{Engine}-Medium has 24 layers and 355 million parameters, on par with GPT2-355M and \textsc{Grover}-Large; (3) \textsc{Engine}-XL has 48 layers and 1.5 billion parameters, on par with GPT2-1.5B and \textsc{Grover}-Mega. In addition, we also show \textsc{Engine} generalizes across architectures by evaluating on LLAMA~\citep{touvron2023llama}. For article retrieval, we implement our method on pretrained vision-language model EVA-CLIP~\citep{sun2023eva}.


\section{Experiments}
\label{sec:experiments}

\subsection{Datasets and Experiment Settings}
\label{sec:experiments_datasets}
\noindent \textbf{Datasets.} We evaluate \textsc{Engine} on three public datasets: GoodNews~\citep{biten2019good}, VisualNews~\citep{liu2020visualnews}, and WikiText~\citep{merity2016pointer}. GoodNews provides the URLs from New York Times from 2010 to 2018. After filtering out broken links or non-English articles, we downloaded 307,286 news articles.  Following the split ratios of~\citet{biten2019good}, we randomly split 15,365 articles for validation, 30,728 articles for testing, and used the rest for training. VisualNews contains news articles from four news sources: \emph{Guardian}, \emph{BBC}, \emph{USA Today}, and \emph{Washington Post}. We obtain 582,194 news articles in total after we removed broken links and articles without metadata. Similarly, we get a 491,796 training set, 28,932 validation set, and a 57,889 test set. WikiText contains 600/60/60 Wikipedia articles in train/test/validation sets, respectively. We performed zero-shot article generation on the test set of WikiText.

\noindent \textbf{Metrics.} Following~\citet{zellers2020defending}, we adopt Perplexity (PPL) \footnote{During inference, we get rid of entity categories from our generated articles to make fair comparisons to other baselines.} to evaluate models on article generation. Perplexity is defined as the exponentiated average negative log-likelihood of a sequence. Given Eq.\ \ref{eq1: lm}, the perplexity of $x$ is calculated by:
\begin{align}
    \text{PPL} (x) = \exp \left \{-\frac{1}{m} \sum^{m}_{i=1}\log p(s_i|s_1,...,s_{i-1}) \right \}
\label{eq5: perplexity}
\end{align}
where $s_1,...,s_{i}$ are ground truth tokens in $x$ and $p(\cdot)$ is the probability predicted by the model. We evaluate models using Recall@K (R@1, R@5, R@10) for article retrieval. 


\smallskip
\noindent \textbf{Implementation Details}
We primarily implemented our models using Pytorch~\citep{paszke2019pytorch} and Transformer~\citep{wolf2020transformers} libraries. The maximum sequence length of language models is set to 1024. For \textsc{Engine}-Base and \textsc{Engine}-Medium, we used a batch size of 8 and a maximum learning rate of $1\times 10^{-4}$. For \textsc{Engine}-XL, we used a batch size of 4 to fit into GPU memory. Correspondingly, the maximum learning rate is set to $2^{0.5}\times10^{-4}$. We finetuned our models for around 3 epochs with 0.06 epoch for linear warm-up on both datasets. We parallelized \textsc{Engine}-XL on 4 NVIDIA RTX-A6000s and \textsc{Engine}-Medium on 2 NVIDIA RTX-A6000s. \textsc{Engine}-XL on VisualNews requires the longest training time- approximately two weeks on our system.

\begin{table*}[t]
\centering
\renewcommand{\arraystretch}{.95}
\setlength{\tabcolsep}{2.5pt}
\begin{tabular}{l c c c c c c}
\toprule
& & & & & GoodNews & VisualNews \\
Model Name & $n_{\rm params}$ & $n_{\rm layers}$ & $d_{\rm model}$ & $n_{\rm heads}$ &  PPL $\downarrow$  & PPL $\downarrow$ \\
\midrule
GPT2-124M~\citep{radford2019language} & 124M & 12 & 768 & 12 & 23.6 & 27.5 \\
\textsc{Grover}-Base~\citep{zellers2020defending} & 124M & 12 & 768 & 12 & 23.8 & 21.9 \\
GPT-Neo-125M~\citep{gao2020pile} & 125M & 12 & 768 & 12 & 27.1 & 29.3 \\
GPT2-124M (Finetuned) & 124M & 12 & 768 & 12 & 17.3 & 18.3 \\
\textsc{Engine}-Base (ClipNE) & 124M & 12 & 768 & 12 & 14.8 & 16.1 \\
\textsc{Engine}-Base (NE) & 124M & 12 & 768 & 12 & \textbf{12.0} & \textbf{13.1} \\

\midrule
GPT2-355M~\citep{radford2019language} & 355M & 24 & 1024 & 16 & 17.8 & 20.1 \\
\textsc{Grover}-Large~\citep{zellers2020defending} & 355M & 24 & 1024 & 16 & 18.5 & 16.4 \\
GPT-Neo-1.3B~\citep{gao2020pile} & 1.3B & 24 & 2048 & 16 & 15.3 & 15.9 \\
GPT2-355M (Finetuned) & 355M & 24 & 1024 & 16 & 13.5 & 14.0 \\
\textsc{Engine}-Medium(ClipNE) & 355M & 24 & 1024 & 16 & 11.6 & 12.5\\
\textsc{Engine}-Medium(NE) & 355M & 24 & 1024 & 16 & \textbf{9.5} & \textbf{10.2}\\

\midrule
GPT2-1.5B~\citep{radford2019language} & 1.5B & 48 & 1600 & 25 & 13.9 & 15.7 \\
\textsc{Grover}-Mega~\citep{zellers2020defending} & 1.5B & 48 & 1600 & 25& 14.5 & 12.6 \\
GPT-Neo-2.7B~\citep{gao2020pile} & 2.7B & 32 & 2560 & 20 & 13.5 & 14.0 \\
GPT-J-6B~\citep{gpt-j} & 6B & 28 & 4096 & 16 & 11.3 & 11.6 \\
GPT2-1.5B (Finetuned) & 1.5B & 48 & 1600 & 25 & 12.6 & 12.4 \\
\textsc{Engine}-XL(ClipNE) & 1.5B & 48 & 1600 & 25 & 10.8 & 11.1\\
\textsc{Engine}-XL(NE) & 1.5B & 48 & 1600 & 25 & \textbf{8.7} & \textbf{9.0}\\

\midrule
LLAMA-7B~\citep{touvron2023llama} & 7B & 32 & 4096 & 32 & 8.3 & 8.5 \\ 
\textsc{Engine}+LLAMA (NE) & 7B & 32 & 4096 & 32 & \textbf{6.5} & \textbf{6.4} \\
\bottomrule
\end{tabular}
\caption{\textbf{Perplexity (PPL) comparison on GoodNews and VisualNews.} ClipNE denotes that we select CLIP-based named entities in \emph{named-entity} field (described in Section~\ref{sec:Engine_NE_extraction}), NE denotes that we apply oracle named entities in \emph{named-entity} field. PPL is calculated only on the article \colorbox{lightmauve}{body}. We observe that our model consistently outperforms baselines of comparable model sizes.} 
\label{table:lm_perplexity}
\end{table*}

\subsection{Article Generation}
\label{sec:experiments_article_generation}

\noindent \textbf{Perplexity.} Table \ref{table:lm_perplexity} presents sizes, architectures, and perplexity results of different models on GoodNews~\citep{biten2019good} and VisualNews~\citep{liu2020visualnews}. We see that \textsc{Engine} variants of all model sizes significantly outperform the baselines. On the base size, \textsc{Engine}-Base(NE) improves PPL over the original GPT2-124M model by a factor of 2 (23.6$\rightarrow$12.0, 27.5 $\rightarrow$ 13.1). We draw three major conclusions from Table \ref{table:lm_perplexity}. First, the data distribution still plays an important role. Finetuned GPT2s improve PPL over the original GPT2s. The improvements become less obvious with a greater model size (VisualNews: 27.5$\rightarrow$18.3 of base size; 15.7$\rightarrow$12.4 of XL size). Second, \textsc{Engine} noticeably improves the performance over finetuned GPTs (4-5 perplexity points on both datasets), which demonstrates the effectiveness of our approach. Third, our contributions are architecture agnostic. \Eg, \textsc{Engine}-LLAMA outperforms LLAMA-7B with an approximately 2-point perplexity improvement on both datasets.

\begin{figure}[t]
    \centering
    \includegraphics[width=0.9\linewidth]{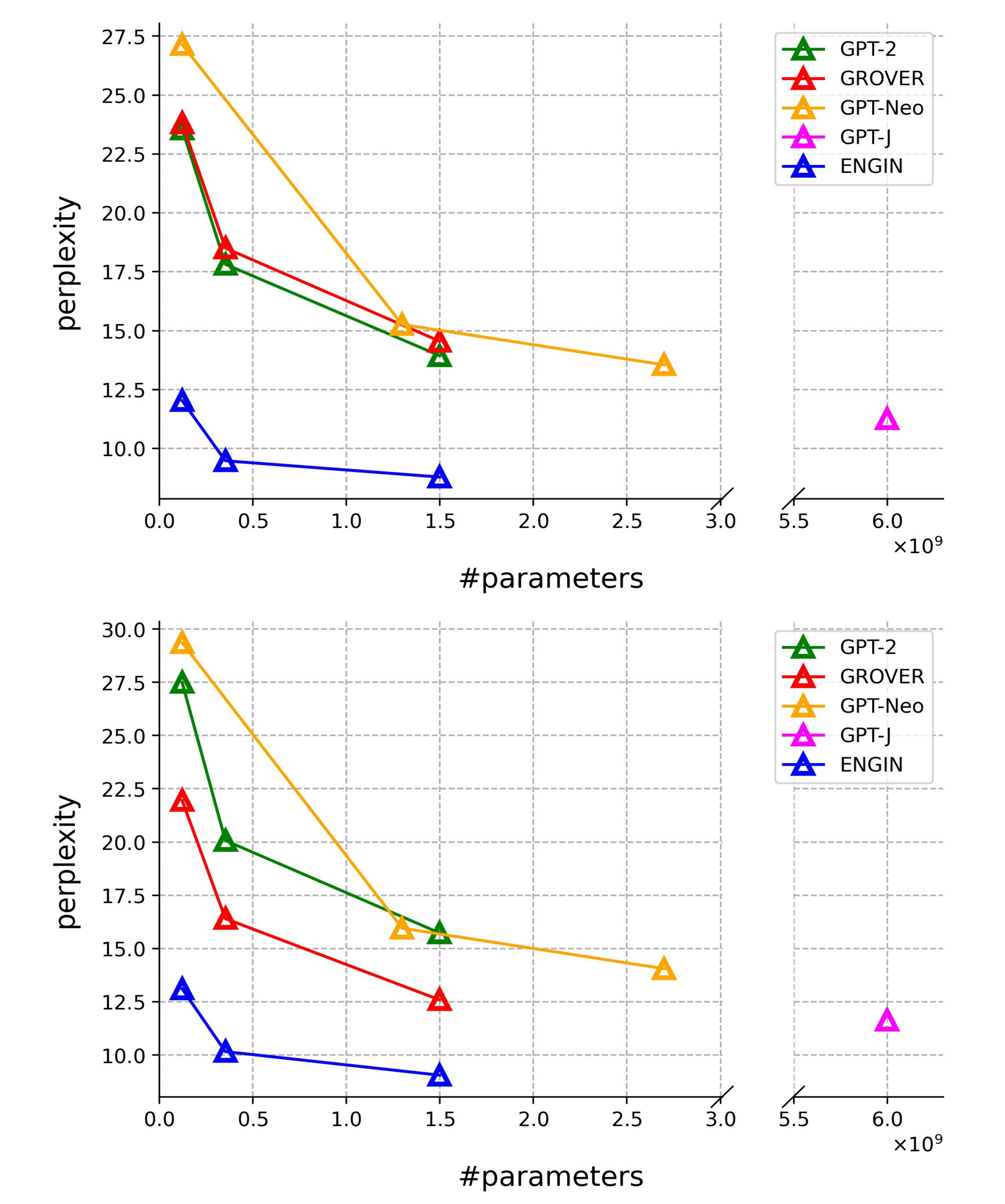}
    \caption{Comparison of the perplexity of different language models on GoodNews (top) and VisualNews (bottom) as a function of learned parameters.}
    \vspace{-2mm}
    \label{fig4: PPL_params}
\end{figure}

\noindent \textbf{Parameter Efficiency.}
Table \ref{table:lm_perplexity} shows that \textsc{Engine} can achieve a comparable performance with alternative models using much fewer parameters. For example, \textsc{Engine}-Base(NE), with only 124M parameters, outperforms the GPT-NEO-2.7B and achieves comparable performance with finetuned GPT2-1.5B (12.0 vs. 12.6 PPL on GoodNews, 13.1 vs. 12.4 PPL on VisualNews). \textsc{Engine}-Medium (NE) model, with 355M parameters, outperforms all the GPT-Series baselines including GPT-J-6B. Figure \ref{fig4: PPL_params} plots the perplexity as a function of the number of parameters on news datasets, demonstrating that \textsc{Engine} gets as good or better results than prior work while also using fewer parameters.

\begin{table}[t]
    \centering
    \setlength{\tabcolsep}{1.5pt}
    \begin{tabular}{rlcc} 
    \toprule
    & & Good- & Visual-\\
    & Model Name & News & News  \\
    \midrule
    \textbf{(A)} & BU~\citep{anderson2018bottomup} & 26.6 & 24.3 \\
    & VisualGLM\citep{du2022glm} & 11.8 & 12.2 \\
    & \textsc{Engine}-1.5B (ours) & \textbf{8.7} & \textbf{9.0} \\
    \midrule
    \textbf{(B)} & InfoSurg~\citep{fung2021infosurgeon} & 41.8 & 42.1 \\
     & InjType~\citep{dong2021injecting} & 18.2 & 19.0 \\ 
     & VNC~\citep{liu2020visualnews} & 16.7 & 17.8 \\
     & \textsc{Engine}-Base (ours) & \textbf{12.0} & \textbf{13.1} \\
    \bottomrule
    \end{tabular}
    \caption{\textbf{PPL of baselines adapted from close-related tasks.} (A) leverages visual features directly extracted from images, (B) contrasts entity-aware mechanisms. Adapting entity-aware mechanisms from related tasks may result in poor performance in our task.}
    \label{table:adapted_baselines}
\end{table}

\noindent \textbf{Additional Baselines adapted from close-related tasks\footnote{InfoSurgeon~\citep{fung2021infosurgeon}, InjType~\citep{dong2021injecting}, VNC~\citep{liu2020visualnews} are adapted from sequence-to-sequence translation, close-ended paragraph generation, and news image captioning, respectively. For a fair comparison, GPT2-Base is applied as the backbone of InjType, VNC, and BU for article generation.}.} In Table \ref{table:adapted_baselines}(A), we see that \textsc{Engine}-1.5B outperforms integrating BU~\citep{anderson2018bottomup} and VisualGLM-6B~\citep{du2022glm} features, demonstrating that our entity-aware mechanism is more effective than directly incorporating image features. As discussed in the Introduction, this is likely due to the loose correlation between images and their corresponding articles. Table \ref{table:adapted_baselines}(B) demonstrates that the entity-aware mechanisms proposed for other text generation tasks do not generalize well to article generation, where our approach obtains a 4-5 PPL improvement on both datasets. We also find that InfoSurgeon struggles to generate articles well, which we argue is due to its sequence-to-sequence translation framework finding it challenging to effectively leverage prior knowledge from pretrained language models. 

\begin{figure}[t]
    \centering
    \includegraphics[width=1.0\linewidth]{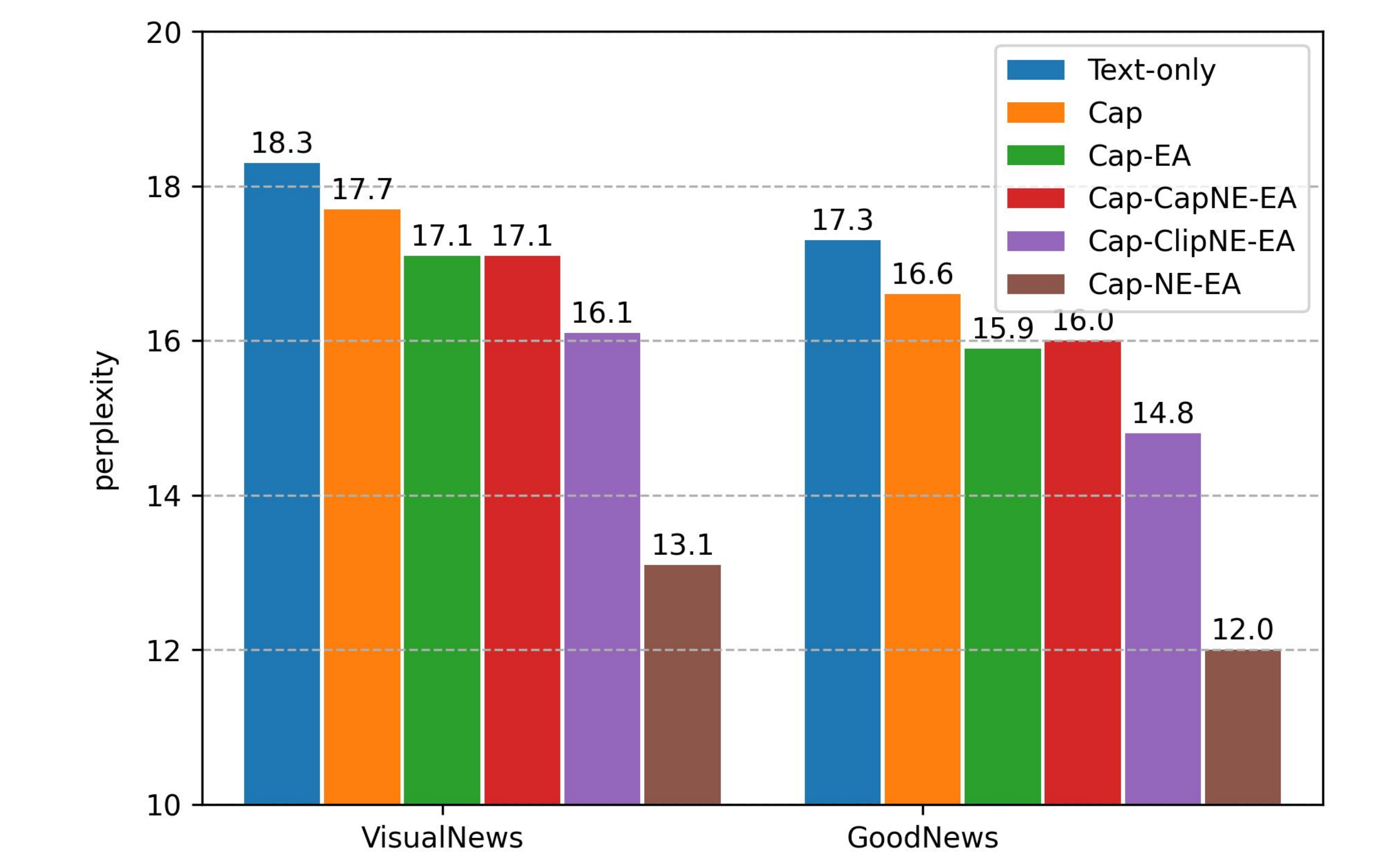}
    \caption{\textbf{Ablation results of \textsc{Engine}-Base.} Text-only denotes the model focuses only on text information, which is same to finetuned GPT2 model. Cap denotes \emph{caption} field, EA denotes the Entity-aware mechanism. CapNE denotes named entities extracted from captions. Both the named-entity extraction module and our entity-aware mechanism can improve the performance.}
    \label{fig:ablation_study}
\end{figure}

\noindent \textbf{Ablation Study.} Figure \ref{fig:ablation_study} shows ablations of \textsc{Engine}-Base. We observe that both the \emph{caption} and \emph{named-entity} fields boost performance, revealing that cues from embedded images help produce higher-quality articles. Comparing using only captions (Cap) vs.\ combining them with our Entity-aware mechanism, we get a minimum gain of 0.6 PPL, demonstrating its effectiveness. In addition, we observe that ClipNE outperforms CapNE, validating that CLIP-detected named entities are more effective than those extracted from captions.

\begin{figure*}[!ht]
    \centering
    \includegraphics[width=.9\linewidth]{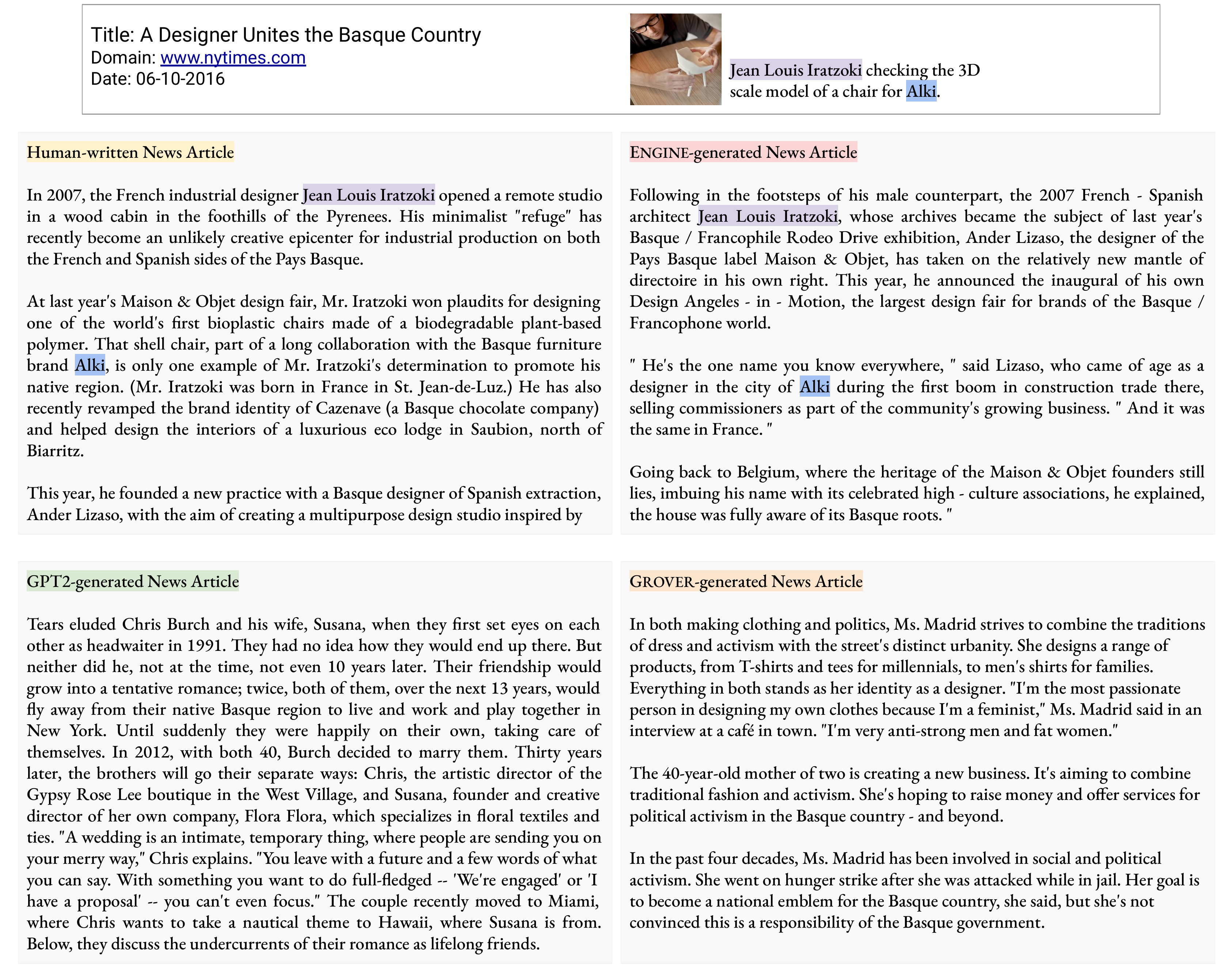}
    \caption{\textbf{Qualitative comparison on GoodNews.} We cut the articles to fit the figure size. The entity names from image information are highlighted in light purple (PERSON tag) and light blue (ORG tag) colors. We can see that the named entities in captions also appear in the human-written and \textsc{Engine}-generated articles. In contrast, the GPT2-generated and \textsc{Grover}-generated articles do not contain correct entity names corresponding to the image.}
     \vspace{-2mm}
    \label{fig:qualitative}
\end{figure*}

\begin{table}[t]
\centering
\begin{tabular}{l rc rc}
\toprule
Method & Human & RoBERTa \\
\midrule
\textsc{Grover}-Mega & 72.8\% & 90\%\\
GPT2-1.5B (Finetuned) & 69.6\% & 84\%\\
\textsc{Engine}-XL & 67.6\% & 84\%\\
\bottomrule
\end{tabular}
\caption{\textbf{Detection of generated articles.} Left column reports the performance of AMT workers at correctly identifying an article as human or machine generated. Right column uses OpenAI's text detector based on RoBERTa~\citep{liu2019roberta} to perform the same task.}
\label{table:human_evaluation}
\end{table}

\begin{table}[t]
\centering
\begin{tabular}{l ccc}
\toprule
Method & 124M & 355M & $\geq$ 1.5B \\
\midrule
GPT2 & 26.1 & 19.1 & 14.8\\
GPT-Neo & 24.9 & \textbf{13.1} & 11.5 \\
GPT-J-6B & - & - & \textbf{9.0} \\
GPT2 (Finetuned) & 33.8 & 25.2 & 25.9\\
\textsc{Engine} (NE) & \textbf{20.7} & 15.4 & 16.3\\
\bottomrule
\end{tabular}
\caption{\textbf{Zero-shot article generation.} We compare the perplexity of \textsc{Engine} to GPT-2~\citep{radford2019language}, GPT-Neo~\cite{gao2020pile}, and GPT-J~\cite{gpt-j} on WikiText~\citep{merity2016pointer}. Both GPT2 and \textsc{Engine} are finetuned on VisualNews~\citep{liu2020visualnews} with a maximum article length of 1024. Though the data distribution between Wikipedia and news is different, our proposed entity-aware method still improves the performance over several baselines.} 
\vspace{-4mm}
\label{table:wikitext_ppl}
\end{table}

\begin{table*}[t]
\centering
\setlength{\tabcolsep}{3pt}
\begin{tabular}{l | ccc|ccc | ccc|ccc}
\toprule
& \multicolumn{6}{c|}{\textbf{Image-to-article}} & \multicolumn{6}{c}{\textbf{Article-to-image}}\\
& \multicolumn{3}{c|}{GoodNews} & \multicolumn{3}{c|}{VisualNews} & \multicolumn{3}{c|}{GoodNews} & \multicolumn{3}{c}{VisualNews} \\
Method & R@1 & R@5 & R@10 & R@1 & R@5 & R@10 & R@1 & R@5 & R@10 & R@1 & R@5 & R@10\\
\midrule
CLIP-B/16 & 19.4 & 42.7 & 52.9 & 30.6 & 53.5 & 64.6 
& 32.5 & 56.8 & 64.4 & 43.1 & 69.5 & 76.5 \\
CLIP-L/14 & 32.5 & 56.0 & 65.3 & 47.3 & 72.1 & 78.9 
& 46.6 & 68.3 & 75.0 & 55.0 & 77.3 & \textbf{84.3} \\
BLIP & 19.8 & 38.0 & 48.8 & 22.0 & 43.6 & 52.5 
& 15.9 & 30.8 & 37.9 & 15.1 & 33.3 & 42.7 \\
EVA01-CLIP-G/14 & 49.6 & 70.3 & 77.1 & 56.9 & 78.9 & 84.7 
& 47.6 & 68.3 & 75.2 & 54.9 & 76.5 & 82.5 \\
EVA02-CLIP-L/14 & 50.7 & 73.1 & 79.4 & 57.9 & 78.6 & 85.2 
& 49.3 & 69.6 & 76.8 & 54.8 & 77.3 & 82.7 \\
\textsc{Engine} (ours) & \textbf{53.8} & \textbf{73.5} & \textbf{79.6} & \textbf{61.9} & \textbf{82.0} & \textbf{86.7} 
& \textbf{51.5} & \textbf{72.0} & \textbf{77.8} & \textbf{56.3} & \textbf{78.3} & 83.9\\ 
\bottomrule
\end{tabular}
\vspace{-2mm}
\caption{\textbf{Retrieval results.} We compare the Recall@K of \textsc{Engine} to CLIP~\citep{radford2021learning}, BLIP~\citep{li2022blip}, and EVA-CLIP~\cite{sun2023eva} on the article-to-image and image-to-article retrieval tasks on GoodNews and VisualNews. For articles with multiple embedded images, we only use the first embedded image. See Section \ref{sec:experiments_article_retrieval} for discussion.}
\vspace{-4mm}
\label{table:article_retrieval}
\end{table*}

\noindent \textbf{Article Quality User Study.} Following~\citep{zellers2020defending, kreps2020all, brown2020language}, we ask annotators to distinguish machine-generated articles from human-written articles. We randomly selected 50 news stories from GoodNews and VisualNews test sets (100 total). Given the metadata and news images, we generated news articles using three different language models: \textsc{Grover}-Mega, GPT2-1.5B (finetuned), and \textsc{Engine}-XL. This results in a total of 200 articles per dataset. We recruited 200 Qualified Amazon Mechanical Turk (AMT) workers per dataset. Each article was annotated 5 times by AMT workers, where each worker was presented with the article titles, images and captions, and was asked to indicate if the article was human or machine generated. If they thought the article was machine-generated, they were asked to indicate a reason for it following the same option format as \citet{tan2020detecting}. Table \ref{table:human_evaluation} reports annotation accuracy in identifying articles from VisualNews as machine or human-generated. We see \textsc{Engine}-XL is able to generate hard-to-detect news articles (a 2\% boost over GPT2-1.5B, and a 5\% gain over \textsc{Grover}-Mega), validating the effectiveness of our approach.

\noindent \textbf{Machine Discriminator.} We apply OpenAI's RoBERTa~\citep{liu2019roberta} detector to detect generated articles. The maximum article length is cut to 512 to fit the model input size. For comparison, we use the same article set from our user study. In Table \ref{table:human_evaluation}, we
observe that the machine discriminator is much better at identifying the machine-generated news. We see that articles produced by \textsc{Engine}-XL can be reliably detected by RoBERTa though it gets the lowest accuracy on human evaluation. This can be due to the fact that \textsc{Grover}-Mega, GPT2-1.5B, and \textsc{Engine}-XL all share a similar underlying model architecture. Thus, they may contain enough similarities in the distributional features that are recognized by the machine discriminator.

\noindent \textbf{Zero-shot Article Generation}
We perform zero-shot experiments on Wikipedia articles to demonstrate \textsc{Engine's} ability to generalize. Table \ref{table:wikitext_ppl} reports performance on WikiText. We find that our oracle named entities and entity-aware mechanism still can improve the performance over several baselines, even though the data distribution between Wikipedia and news is significantly different. For example, the GPT2 models finetuned on Visualnews get worse performance than the original GPT2 models on WikiText. However, our \textsc{Engine} models get comparable or better results than the original GPT2 across different model sizes.

\noindent \textbf{Qualitative Results.} We provide a qualitative comparison of GoodNews articles in Figure \ref{fig:qualitative}. Consistent with our annotation experiment, we compare the human-written article with three machine-generated articles. From the results, we can see that \textsc{Engine}-XL can effectively produce articles with the named entities learned from image information. In contrast, finetuned GPT2-1.5B and \textsc{Grover}-Mega failed to generate correct named entities in articles. For example, both \textsc{Engine}-generated article and the human-written article mentioned ``Hean Louis Iratzoki'' and ``Alki'', which are appeared in the caption. In contrast, articles generated by GPT2 or \textsc{Grover} are discussing some other entities such as ``Chris Burch'' and ``Ms. Madrid.'' 

\subsection{Article Retrieval}
\label{sec:experiments_article_retrieval}
Table \ref{table:article_retrieval} compares the Recall@K retrieval scores of \textsc{Engine} with the state-of-the-art baselines on the test splits of GoodNews and VisualNews. Following~\citet{tan2022newsstories}, we randomly select 1500 article-image pairs from each dataset for evaluation. In the image-to-article retrieval task, we observe that compare to EVA02-CLIP-L/14~\citep{sun2023eva}, \textsc{Engine} boost the Recall@1 scores from 50.7 to 53.8 on GoodNews and from 57.9 to 61.9 to VisualNews. For article-to-image retrieval, \textsc{Engine} achieves the highest performance with Recall@1 scores of 51.5 and 56.3 on GoodNews and VisualNews, respectively. The retrieval results validate the importance of named entities within articles and the effectiveness of our proposed method.

\section{Discussion}
In our paper, we mainly investigate modeling machine-generated articles, which can be used directly for generation while also can providing strong language features to support applications like article retrieval. However, actors can also use the same technology to generate articles for misinformation by modifying information of specific fields to realize two potential purposes: monetization (ad revenue through clicks) or propaganda (communicating targeted information)~\citep{zellers2020defending}. Thus, the development of a better article generator can not only help humans write high-quality articles but also potentially help train a more powerful discriminator. Table~\ref{table:human_evaluation} reports the performance of using human judgements or OpenAI's RoBERTa-based machine generated text detector~\citep{liu2019roberta}. When comparing the results of the RoBERTa detector for GPT2-1.5B and \textsc{Engine}-XL, we find that only 25\% of the articles that were predicted as human written came from the same generation prompts. Thus, the two methods can provide different views of the same prompt, which can provide additional information for training an even more powerful machine generated text detector. We note that our contributions are largely architecture agnostic, so they could also be used in RNN-based generators, which may provide a larger distribution shift in the generated articles that may fool a discriminator trained only on Transformer outputs.

\section{Conclusion}
In this paper, we proposed \textsc{Engine}, an entity-aware article generation and retrieval method that explicitly incorporates and models named entities in language models. Concretely, \textsc{Engine} extracts named entities from both metadata and embedded images in articles, providing a more comprehensive source of entity information. In addition, we introduce an entity-aware mechanism to help \textsc{Engine} recognize and predict named entities more effectively and accurately. \textsc{Engine} outperforms current popular language models in quantitative and qualitative experiments on GoodNews, VisualNews, and WikiText. For example, \textsc{Engine} outperforms GPT-J by roughly 2.5 perplexity points using only a quarter parameters of GPT-J and boost the performance of EVA02-CLIP by 3-4 Recall@1 accuracy in article retrieval experiments. The noticeable improvements demonstrate that \textsc{Engine} can generate and retrieve articles more accurately and efficiently by effectively leveraging named entities. 

\noindent\textbf{Acknowledgements}
This material is based upon work supported, in part, by DARPA under agreement number HR00112020054. Any opinions, findings, and conclusions or recommendations expressed in this material are those of the author(s) and do not necessarily reflect the views of the supporting agencies.

\section*{Limitations}
We discuss limitations and potential improvements of our work in this section. First, though our method can effectively predict the correct entity names in articles, their corresponding entity categories might be mistakenly predicted. For example, in Figure \ref{fig:qualitative}, the brand name ``Alki'' is recognized as a city name by \textsc{Engine}. Therefore, a more accurate entity-aware mechanism could be developed in future work. Second, the image information can be further explored. In this paper, we mainly investigate the captions and named entities of news images. However, other information such as the locations of images within articles may also prove useful for article generation. In addition, our current methods detect named entities from images considering each entity independently using a text-image matching framework. However, since the relationships between entities also affect the probability that entities appear in images, the incorporation of entity relationships can also be considered to further improve the entity detection module.

\section*{Ethics Statement}
\textsc{Engine} is a model for article generation and retrieval. It can either help automated journalism or defending against machine-generation articles. However, there is no perfect system which can generate 100\% accurate articles. Therefore, it is critical for practitioners to check the fact mentioned in articles and avoid the misinformation brought by failure generation cases. Additionally, someone could use our approach to generate misinformation.  However, \textsc{Engine} applies the network structure that is same to GPT2, which means the discriminators trained for GPT2 articles (\eg, RoBERTa detector~\citep{liu2019roberta}) are also effective to discriminate \textsc{Engine}-generated articles.  Our paper helps to highlight the need for building tools like the RoBERTa detector.


\bibliography{anthology,custom}
\bibliographystyle{acl_natbib}

\appendix
\section{Additional Experiment Results}
\subsection{Article Generation}
\begin{figure*}[t]
\begin{center}
  \includegraphics[width=1\linewidth]{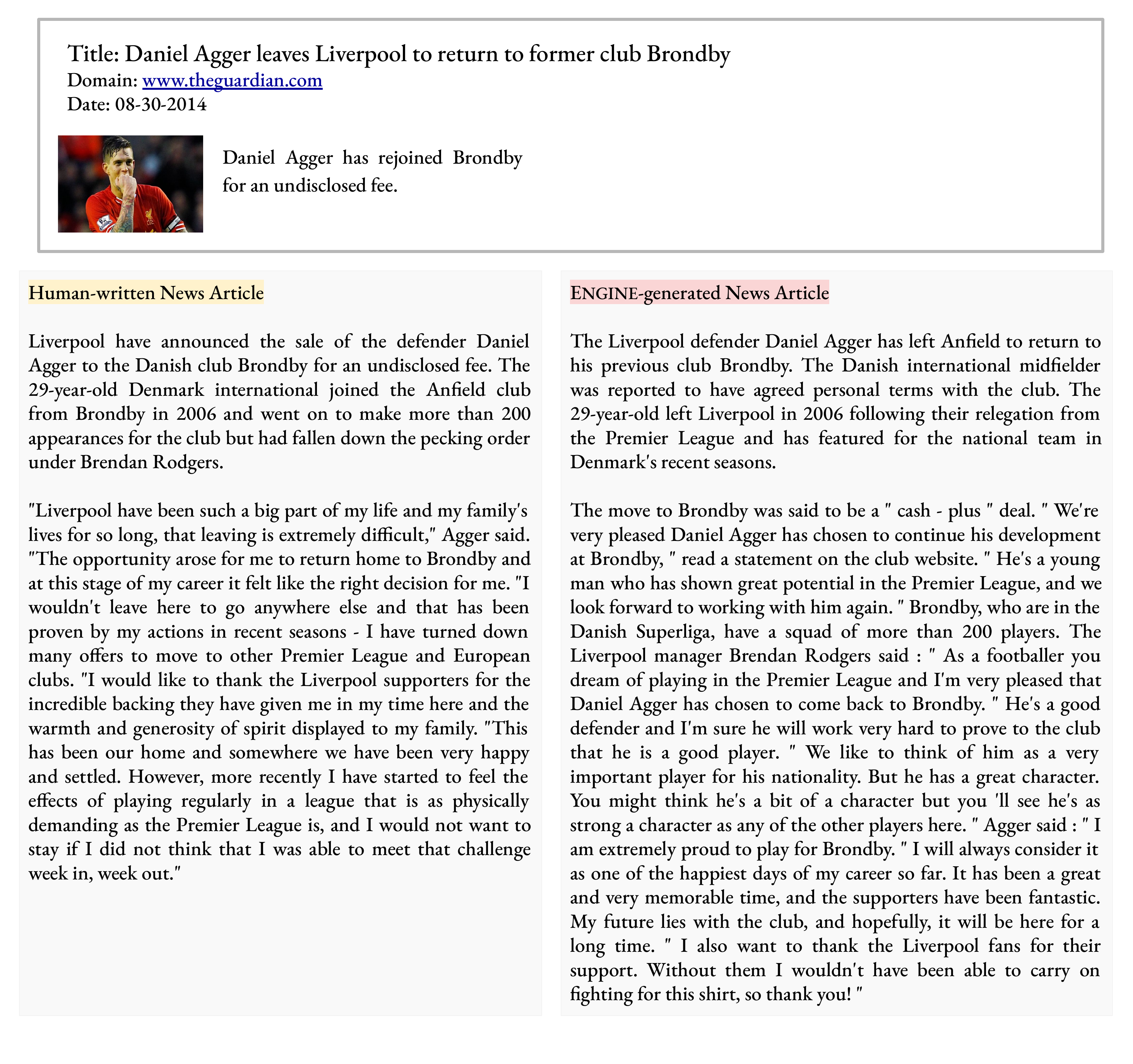}
\end{center}
  \caption{Example of human-written and \textsc{Engine}-generated news articles from \emph{Guardian}.
  }
\label{fig: visualnews_guardian}
\end{figure*}

\begin{figure*}[t]
\begin{center}
  \includegraphics[width=1\linewidth]{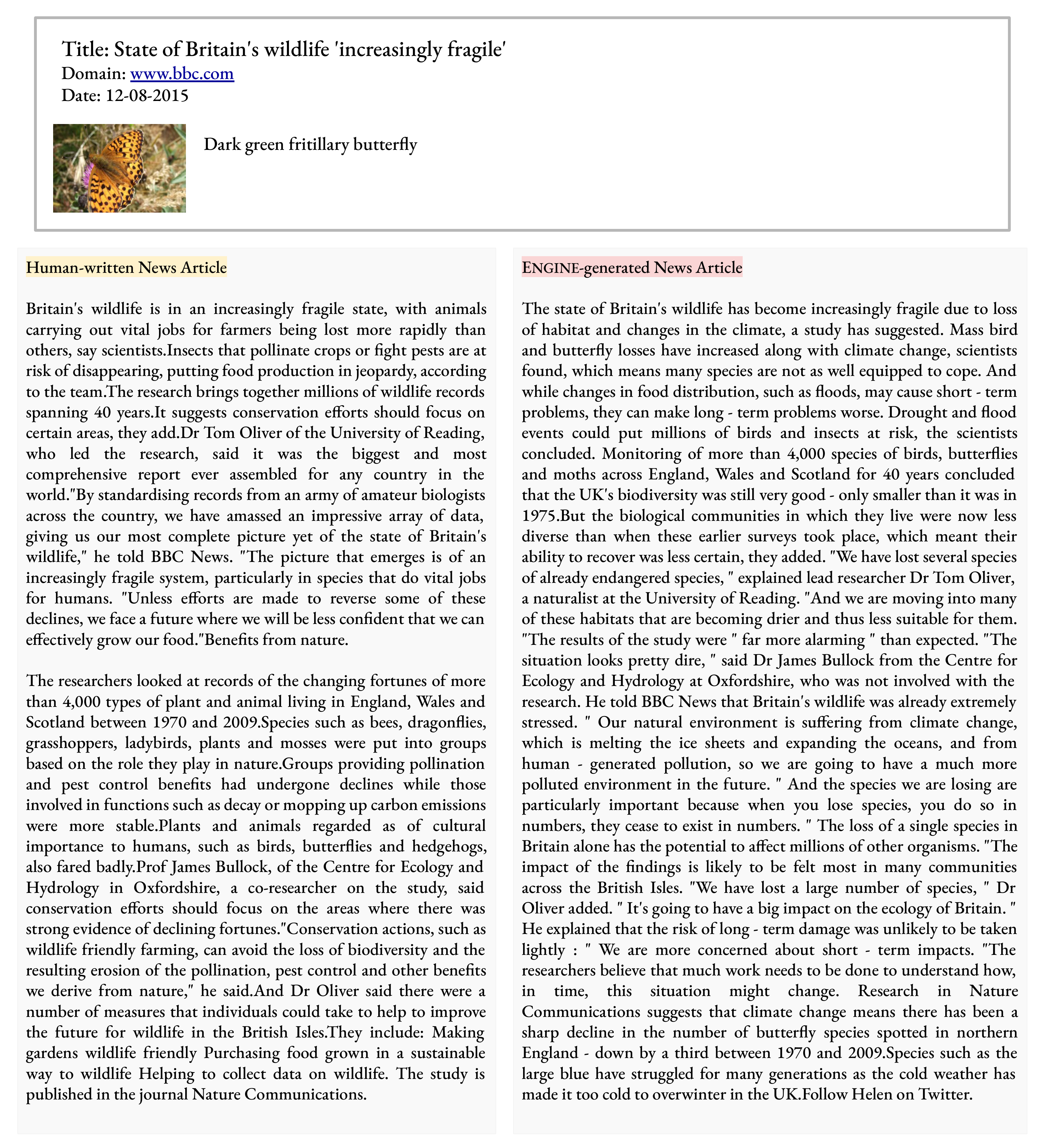}
\end{center}
  \caption{Example of human-written and \textsc{Engine}-generated news articles from \emph{BBC}.
  }
\label{fig: visualnews_bbc}
\end{figure*}

\begin{figure*}[t]
  \centering
  \includegraphics[width=1.0\linewidth]{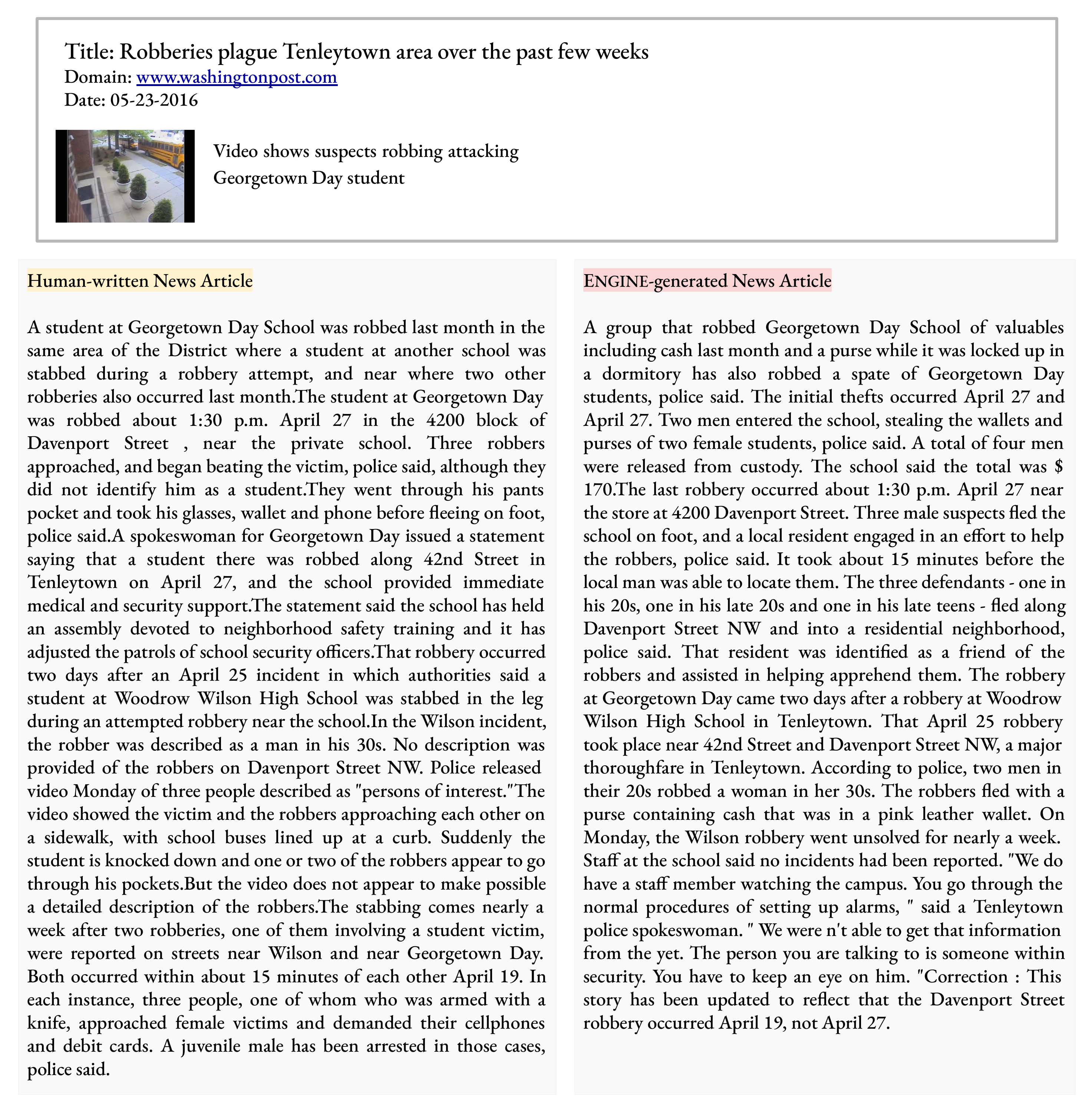}
  \caption{Example of human-written and \textsc{Engine}-generated news articles from \emph{Washington Post}.
  }
\label{fig: visualnews_post}
\end{figure*}

\begin{figure*}[t]
\begin{center}
  \includegraphics[width=1\linewidth]{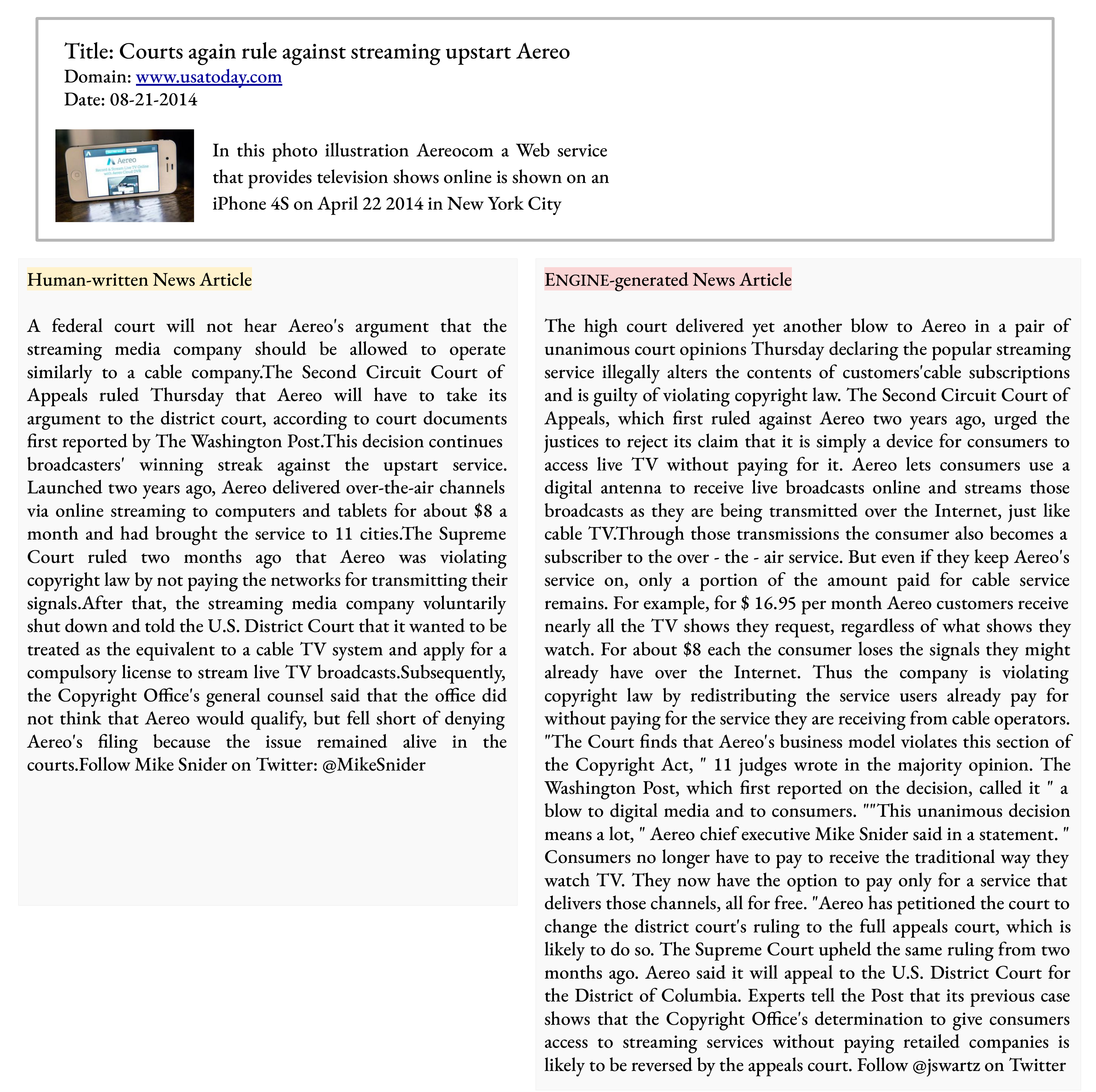}
\end{center}
  \caption{Example of human-written and \textsc{Engine}-generated news articles from \emph{USA Today}.
  }
\label{fig: visualnews_usatoday}
\end{figure*}

\begin{figure*}[t]
\begin{center}
  \includegraphics[width=1\linewidth]{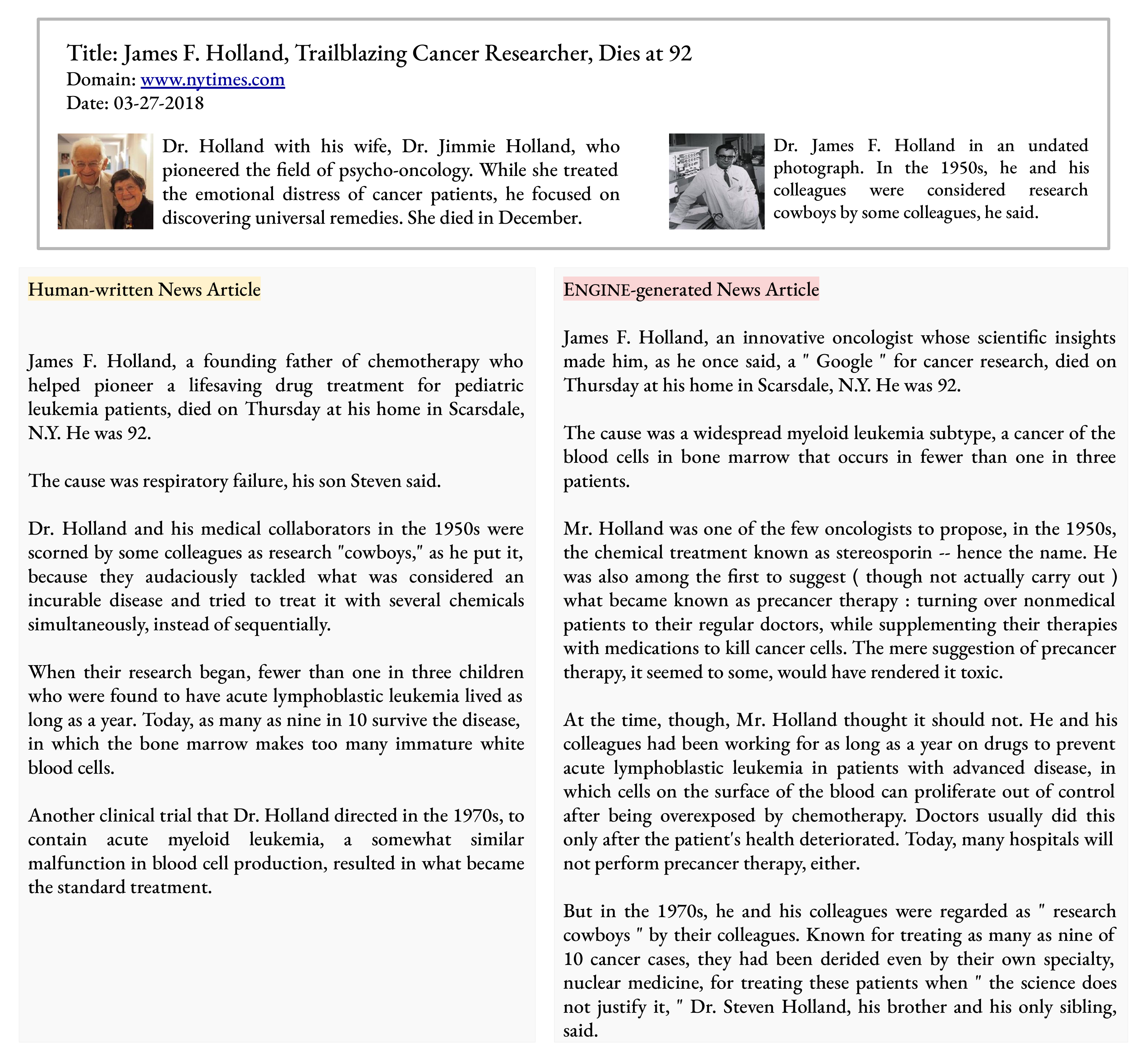}
\end{center}
  \caption{Example of human-written and \textsc{Engine}-generated news articles from \emph{New York Times}.
  }
\label{fig: goodnews_nytimes}
\end{figure*}

\begin{figure*}[t]
\begin{center}
  \includegraphics[width=1\linewidth]{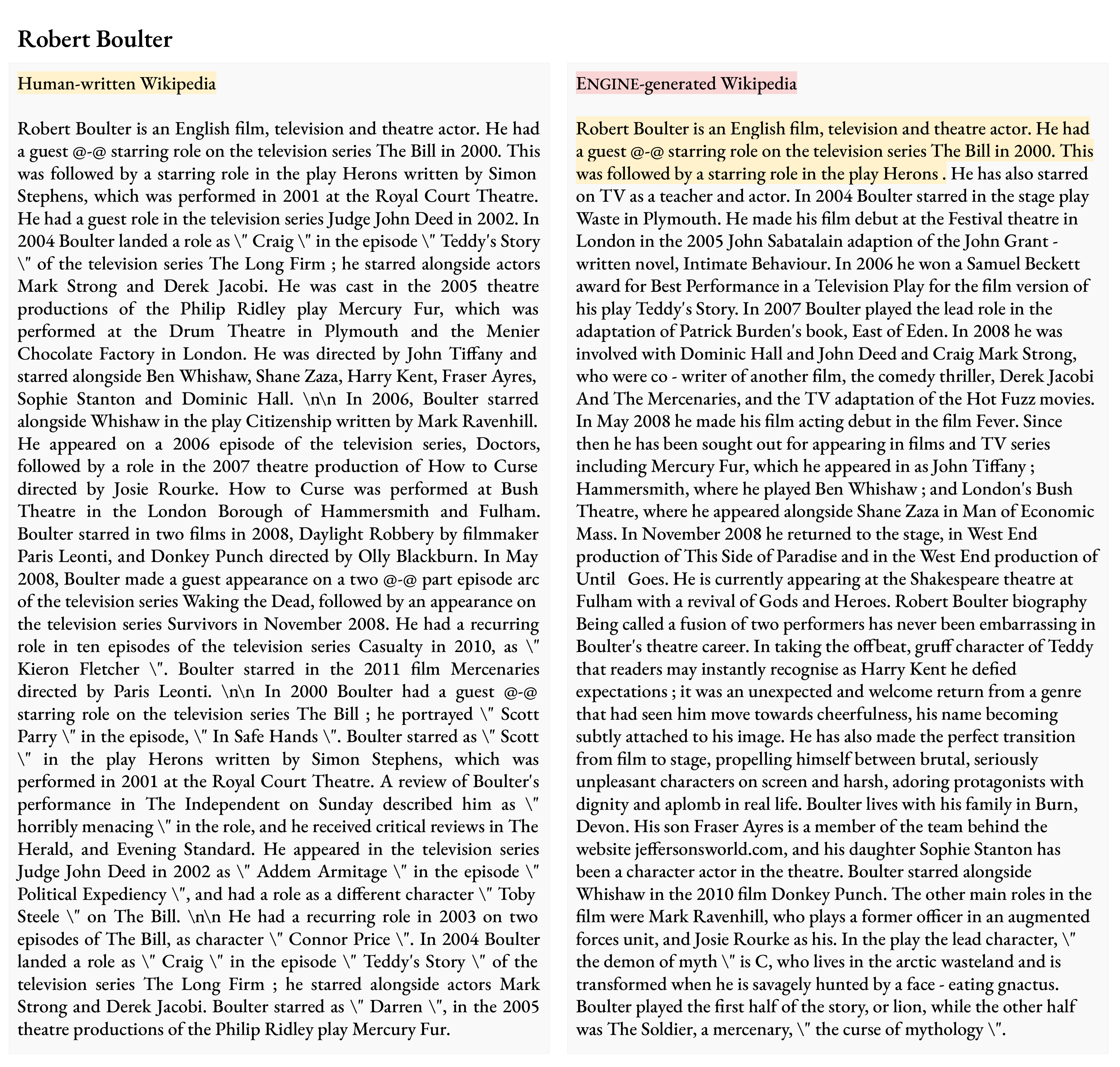}
\end{center}
  \caption{Example of human-written and \textsc{Engine}-generated Wikipedia articles (from WikiText dataset~\citep{merity2016pointer}). We generate articles conditioned on both the title \textbf{Robert Boulter}, and the first 50 tokens highlighted by light yellow.  The articles are cut to fit the figure size.
  }
\label{fig: wiki_robert}
\end{figure*}

\begin{figure*}[t]
\begin{center}
  \includegraphics[width=1\linewidth]{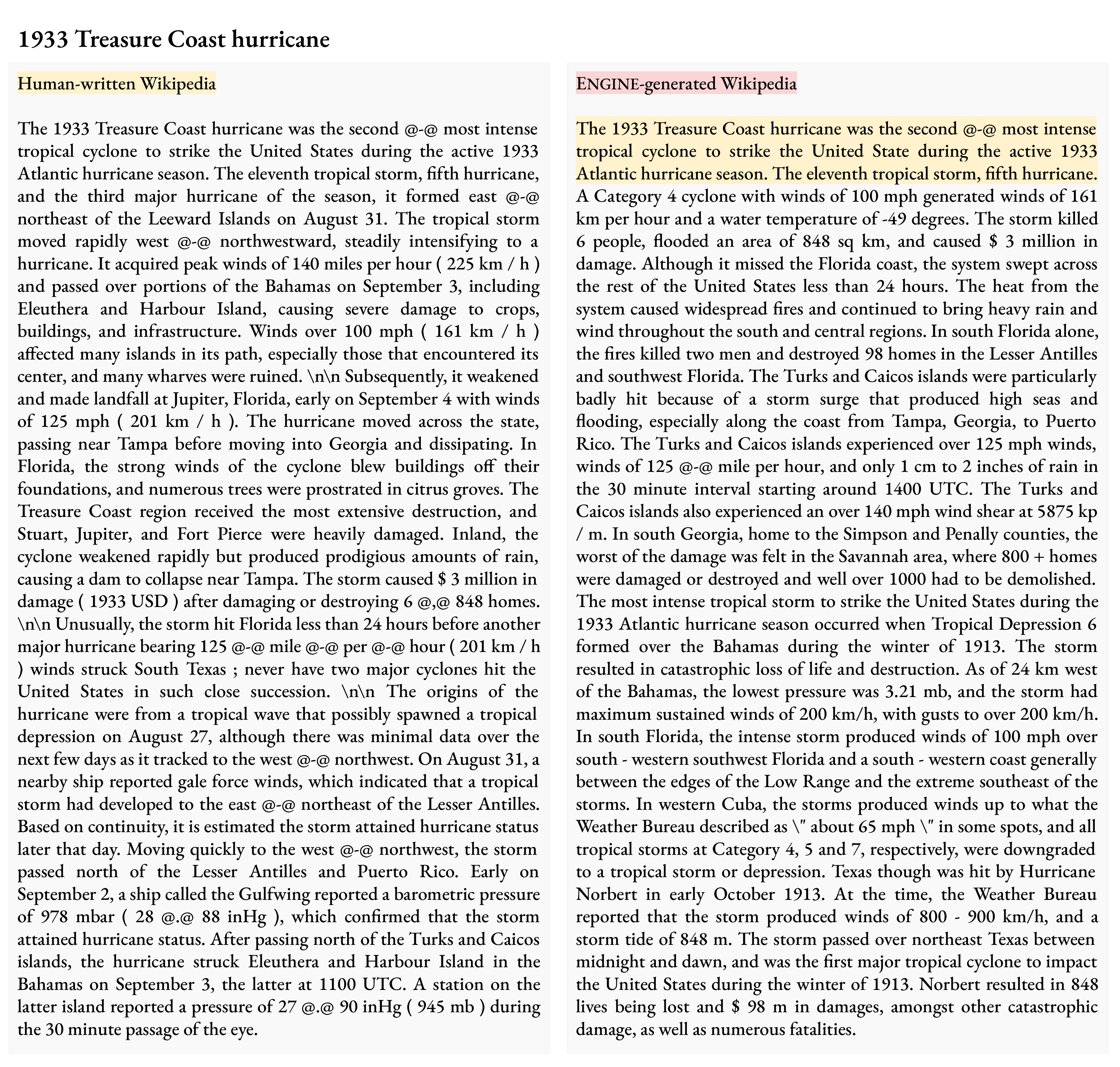}
\end{center}
  \caption{Example of human-written and \textsc{Engine}-generated Wikipedia articles about \textbf{1993 Treasure Coast hurricane}. 
  }
\label{fig: wiki_hurricane}
\end{figure*}

Additional qualitative results on article generation are provided in Figure \ref{fig: visualnews_guardian}, \ref{fig: visualnews_bbc}, \ref{fig: visualnews_post}, \ref{fig: visualnews_usatoday}, \ref{fig: goodnews_nytimes}, \ref{fig: wiki_robert}, \ref{fig: wiki_hurricane} as the supplement of our main paper, demonstrating that \textsc{Engine} is able to generate articles given different news sources.

\subsection{Article Quality Annotation}
Table \ref{tab2: human_evaluation_appendix} reports human accuracy in identifying articles as machine or human-generated to supplement the main paper. We see that \textsc{Engine}-XL is able to generate more realistic-looking news articles on Goodnews, consistent with the conclusion in our main paper.

\begin{table}[ht]
\centering
\begin{tabular}{rl c}
\toprule
& \multicolumn{2}{l}{\textbf{Human-based detector}} \\
\midrule

& \textsc{Grover}-Mega & 76.4\% \\ 
& GPT2-1.5B (Finetuned) & 73.6\% \\ 
& \textsc{Engine}-XL & \textbf{70.4\%} \\ 
\bottomrule
\end{tabular}
\vspace{-2mm}
\caption{The performance of AMT workers at correctly identifying an article as human or machine generated on GoodNews.}
\label{tab2: human_evaluation_appendix}
\vspace{-4mm}
\end{table}

\subsection{Ablation Study on Top-k Named Entities}
We provide the ablation study on top-k named entities in Table \ref{tab: ppl_vs_topk}. We see that the model achieves the best performance when $k$ is set to 15. When $k$ is greater than 10, the improvement is limited.

\begin{table}[!ht]
    \centering
    \setlength{\tabcolsep}{3.pt}
    \begin{tabular}{lcccc}
    \toprule
        top-k named entities & 5 & 10 & 15 & 20 \\
    \midrule
        PPL $\downarrow$ & 15.5 & 14.8 & 14.5 & 14.6 \\
    \bottomrule
    \end{tabular}
    \caption{Ablation study on the number of named entities detected by CLIP (GoodNews).}
    \label{tab: ppl_vs_topk}
\end{table}

\subsection{Recall of CLIP-detected Named Entities}
We use Oracle NE as ground truth labels to evaluate the retrieval results by CLIP model. In Table \ref{tab: recall_clipne}, we observe that approximately 30\% named entities from Oracle NE have been retrieved by CLIP. In contrast, recall of Cap on GoodNews and VisualNews are 22.57\% and 7.41\% respectively. The gap between GoodNews and VisualNews is likely because captions in GoodNews are often much longer than captions in VisualNews. The retrieval results validate that ClipNE contains more relevant named entities compared to named entities extracted solely based on captions.

\begin{table}[!ht]
    \centering
    \begin{tabular}{l cc}
    \toprule
        Named Entities & GoodNews & VisualNews \\
        \hline
        Cap & 22.57 & 7.41 \\
        ClipNE & 29.84 & 31.19 \\
    \bottomrule
    \end{tabular}
    \vspace{-2mm}
    \caption{Recall of Cap, ClipNE on GoodNews and VisualNews. Cap represents named entities that appear in captions, ClipNE represent named entities detected by CLIP model.}
    \vspace{-4mm}
    \label{tab: recall_clipne}
\end{table}

\subsection{Ablation Study on the Canonical Order}
The ablation study of varying inference order is shown in Table \ref{tab:ablation_canonical_order}. From the Table, we see that canonical orders which are not consistent with the training order result in greater PPL of the language model. The model achieves the best performance when the inference order is aligned with the training order.

\begin{table}[!ht]
    \centering
    \begin{tabular}{l c}
    \toprule
        canonical order & GoodNews PPL$\downarrow$ \\
        \hline
        date-domain-title-summary & 18.2 \\
        title-date-domain-summary & 19.2 \\
        summary-date-domain-title & 20.5 \\
        domain-date-title-summary & 17.3\\
    \bottomrule
    \end{tabular}
    \caption{Ablation study on the canonical order during inference (GoodNews).}
    \label{tab:ablation_canonical_order}
\end{table}


\subsection{Ablation Study on Article Retrieval}
We ablate the two modules of our method on article retrieval in Table \ref{table:ablation_article_retrieval}. Text-only denotes that we directly use the original articles as the input. NE denotes named entity extraction, and EA denotes the entity-aware mechanism. From the table, we observe that both named entity extraction and the entity-aware mechanism can boost the article retrieval performance. 


\begin{table}[ht]
\centering
\setlength{\tabcolsep}{3pt}
\begin{tabular}{l | ccc | ccc}
\toprule
& \multicolumn{3}{c|}{\textbf{Image-to-article}} & \multicolumn{3}{c}{\textbf{Article-to-image}} \\
Method & R@1 & R@5 & R@10 & R@1 & R@5 & R@10 \\
\midrule
Text-only & 57.9 & 78.6 & 85.2 & 54.8 & 77.3 & 82.7 \\ 
NE & 60.0 & 80.3 & 85.9 & 56.1 & 78.2 & 83.7 \\ 
NE-EA & \textbf{61.9} & \textbf{82.0} & \textbf{86.7} & \textbf{56.3} & \textbf{78.3} & \textbf{83.9} \\ 
\bottomrule
\end{tabular}
\caption{Ablation study on article retrieval (VisualNews). Both named entity extraction and the entity-aware mechanism can improve the article retrieval performance on VisualNews. }
\vspace{-4mm}
\label{table:ablation_article_retrieval}
\end{table}

\subsection{Decoding Strategy for Article Generation} 
Likelihood-maximization decoding strategies like greedy search or beam search work well in close-ended generation such as image captions, machine translation, or summarization. However, these methods suffer from the repetitive text problem in open-ended generations like dialog or story generation~\citep{hashimoto2019unifying, holtzman2019curious}. Sampling methods~\citep{fan2018hierarchical, holtzman2019curious} are therefore proposed to introduce more randomness and surprise to text generation. In our work, we adopt the top-p sampling (nucleus sampling) method~\citep{holtzman2019curious} as our decoding strategy.


\subsection{Article Quality Annotation Templates}
\begin{figure*}[t]
\begin{center}
  \includegraphics[width=1\linewidth]{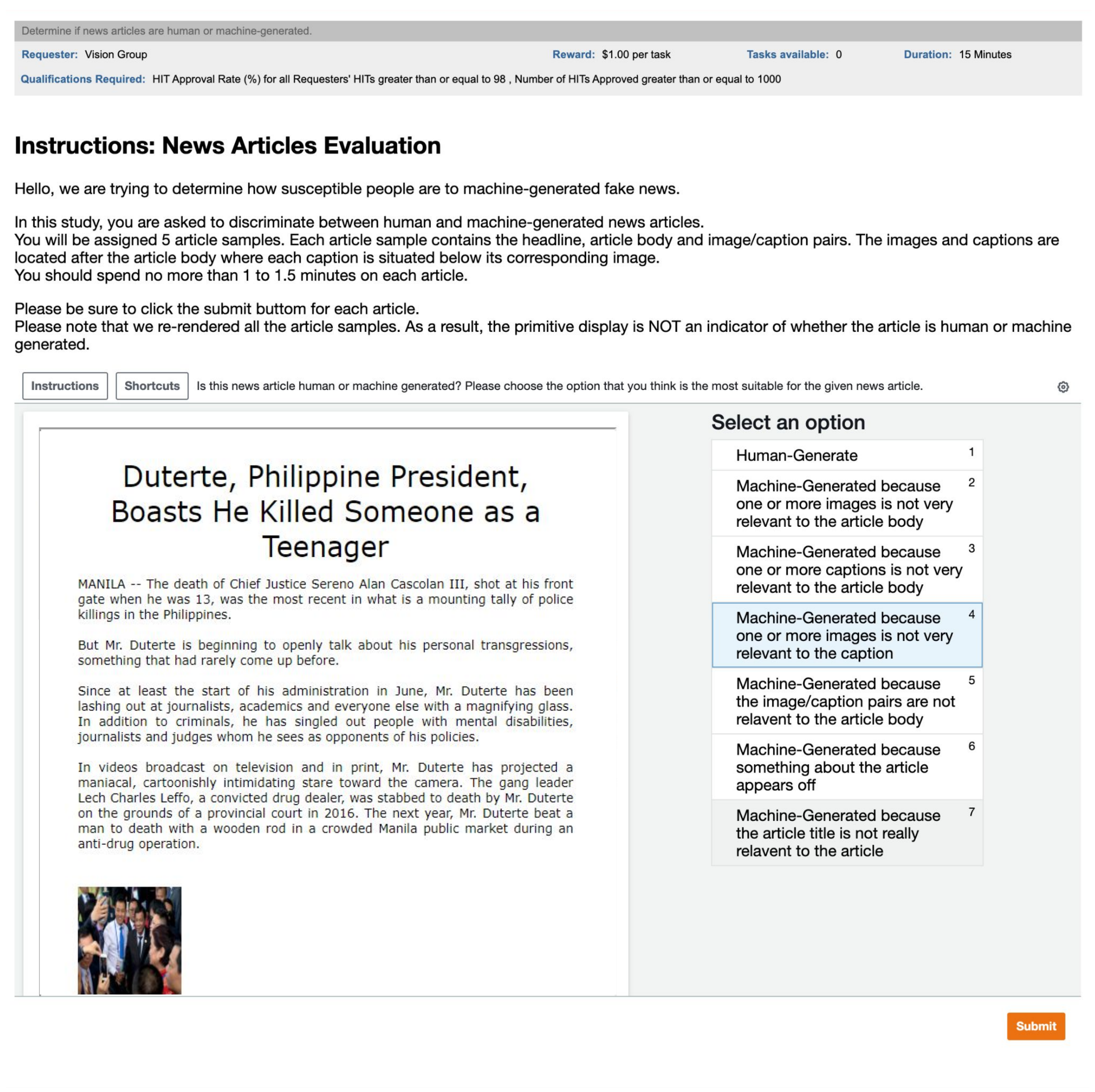}
\end{center}
  \caption{The interface used by AMT workers in our article quality annotation experiment.
  }
\label{fig: user_interface}
\end{figure*}

Following~\citep{tan2020detecting}, annotators are asked to indicate a reason for whether the articles are human-written or machine-manipulated. We provide a view of the AMT worker interface in Figure \ref{fig: user_interface}.

\clearpage

\end{document}